\documentclass{article} 
\usepackage{iclr2027_conference,times}

\usepackage{amsmath,amsfonts,bm}
\usepackage{colortbl}

\def\eqref#1{equation~\ref{#1}}

\def\1{\bm{1}}

\DeclareMathAlphabet{\mathsfit}{\encodingdefault}{\sfdefault}{m}{sl}
\SetMathAlphabet{\mathsfit}{bold}{\encodingdefault}{\sfdefault}{bx}{n}

\usepackage{hyperref}
\usepackage{url}

\usepackage{microtype,graphicx,subcaption,booktabs,multirow}
\usepackage{mathtools}
\usepackage[capitalize,noabbrev]{cleveref}
\usepackage{tcolorbox}
\tcbuselibrary{listings,breakable}

\title{Latent Chain-of-Thought as Planning: Decoupling Reasoning from Verbalization}

\author{Jiecong Wang \\
Beihang University \\
\texttt{jcwang@buaa.edu.cn}
\And
Hao Peng \\
Beihang University \\
\texttt{penghao@buaa.edu.cn}
\And
Zhanyi Wang \\
QI-ANXIN GROUP \\
\texttt{wangzhanyi@qianxin.com}
\And
Chunyang Liu \\
Didi International Business Group \\
\texttt{liuchunyang@didiglobal.com}
\And
Guanlin Wu \\
National University of Defense Technology \\
}

\iclrfinalcopy
\begin{document}
\maketitle
\fancyhead{}
\renewcommand{\headrulewidth}{0pt}

\begin{abstract}
Chain-of-Thought (CoT) empowers Large Language Models (LLMs) to tackle complex problems, but remains constrained by the computational cost and early token commitments in discrete reasoning traces. 
Recent latent reasoning approaches attempt to optimize efficiency by performing reasoning within continuous hidden states. 
However, many such methods optimize latent states end to end without a trained interface for intermediate textual readout, and several representative configurations use a pre-defined number of latent steps during inference.
In this work, we introduce \textbf{PLaT} (\textbf{P}lanning with \textbf{La}tent \textbf{T}houghts), a framework that decouples latent planning from verbalization.
The Planner deterministically evolves latent planning states, while an independent Decoder provides textual readouts when needed.
Answer-aware textual stopping allows the latent rollout to use a problem-dependent number of groups rather than a pre-specified chain length.
PLaT achieves competitive coverage at larger $k$ in several mathematical settings, with lower Pass@1: on Llama-1B GSM8K, it reaches 80.59\% Pass@128 versus CODI's 72.37\%. These results support PLaT as a candidate-generation interface supplying multiple textual readouts for downstream verification or reranking.
\end{abstract}

\section{Introduction}

\begin{figure}[ht]
  \centering
  \includegraphics[width=0.88\linewidth]{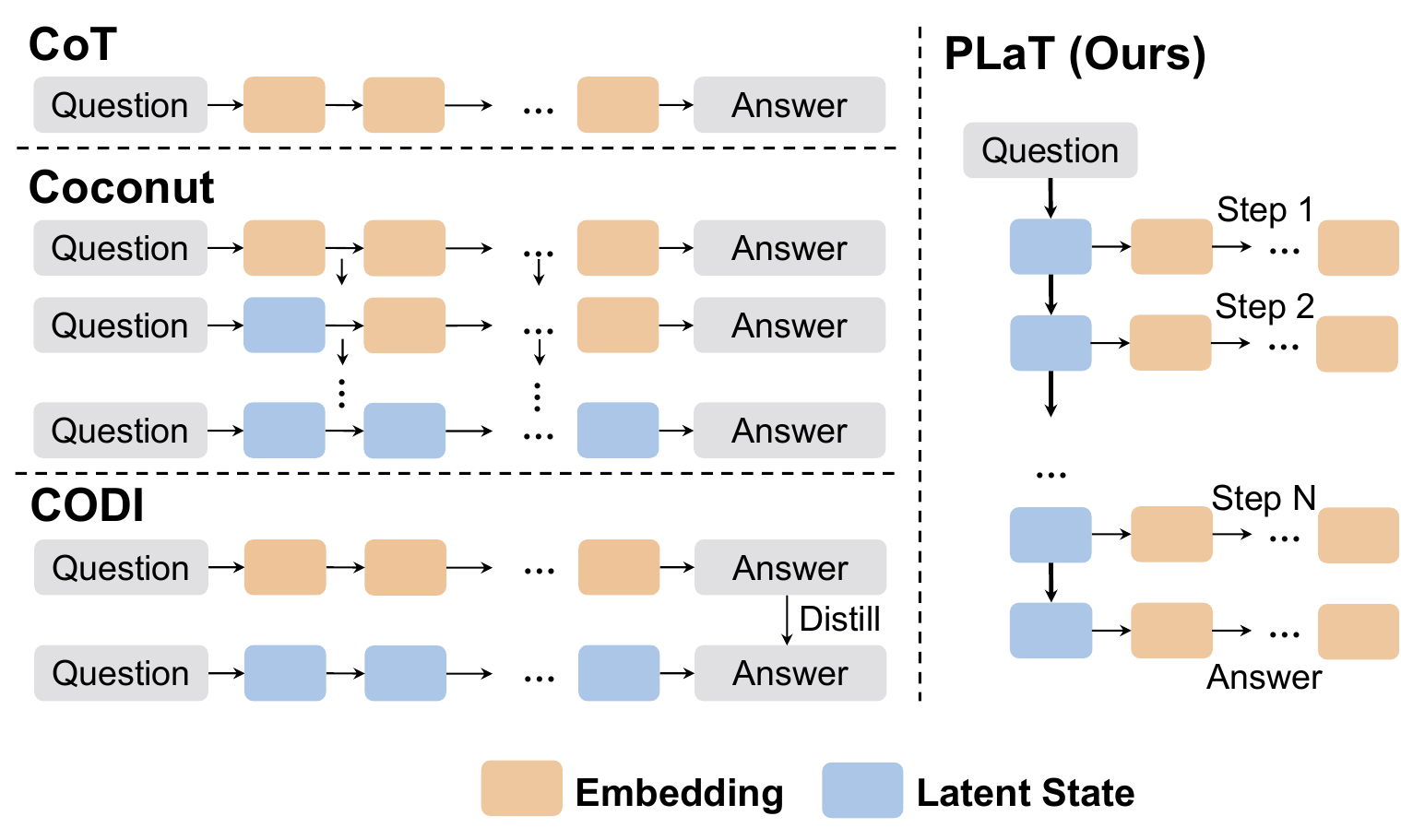}
  \caption{\textbf{Comparison of PLaT and other reasoning strategies.} Orange blocks denote token embeddings and blue blocks denote continuous latent states. CoT generates explicit reasoning tokens; Coconut progressively replaces textual reasoning with latent states during training; CODI distills explicit reasoning into a latent chain. PLaT separates latent-state evolution from textual readout: the vertical chain represents planning, and each horizontal branch represents a decoded reasoning segment or answer. Decoded text is not fed back into the Planner.}
  \label{fig:intro-compare}
\end{figure}

Chain-of-Thought (CoT) reasoning \cite{wei2022chain,kojima2022large,wang2023selfconsistency,zhang2023automatic} improves the reasoning performance of Large Language Models (LLMs) by decomposing problems into sequences of intermediate steps \cite{zhouleast}. 
This approach supports applications ranging from code generation \cite{chen2021evaluating, chen2023program,li2025structured,liu2025revisiting} to autonomous agents \cite{yao2022react,shinn2023reflexion,schick2023toolformer}.
However, each sampled trajectory commits to discrete textual choices.
At every generation step, a sampled trajectory selects one discrete token, while alternative valid continuations remain available only across separate samples \cite{yao2023tree,zhang2025soft,chen2025reasoning}. 
This can make recovery from an incorrect intermediate choice difficult within a single sampled trajectory.
Additionally, generating long sequences of intermediate tokens incurs substantial inference cost \cite{zhang2025soft,sui2025stop,wang2025system15reasoningtraversallanguage,feng2025efficient}.

To mitigate the inefficiency, recent works have explored latent reasoning, where the model evolves hidden states internally before outputting a final answer \cite{zhang2025soft,xu2025softcot,hao2025training,shen2025codi,tan2025think}. 
While promising, they predominantly adopt an end-to-end implicit paradigm, optimizing latent states directly for the final generation. 
This leaves two practical gaps that motivate our design. 
First, intermediate states in many latent reasoning methods lack a trained interface for textual inspection.
Second, several prominent methods, including the Coconut and CODI configurations evaluated here, use a prespecified number of latent steps during inference.
This fixes their latent rollout budget regardless of problem difficulty, motivating an architecture that permits a problem-dependent chain length \cite{li2025system}.

We argue that a robust reasoning system should mirror the cognitive distinction between thought and language. 
From a cognitive perspective, language serves merely as a low-dimensional projection (interface) of high-dimensional thought; the core reasoning process often occurs implicitly without verbalization \cite{varley2000evidence,fedorenko2016language,coetzee2022dissociating,fedorenko2024language}. 
Ideally, a continuous state can support multiple possible textual continuations until an interface with the external world is required. In PLaT, this is expressed as a distribution over Decoder readouts; the deterministic Planner itself does not sample multiple latent trajectories.
Motivated to replicate this implicit process computationally, we draw inspiration from Multi-Token Prediction (MTP) \cite{stern2018blockwise,qi2020prophetnet,gloeckle2024better,nagarajan2025roll,samragh2025your} (that transformer hidden states are able to encode information about future tokens before they are generated) and propose to model reasoning as a sequence of latent planning states.

In this work, we introduce \textbf{PLaT} (\textbf{P}lanning with \textbf{La}tent \textbf{T}houghts), a framework that decouples the reasoning process from verbalization. 
As illustrated in Figure~\ref{fig:intro-compare}, our architecture comprises two distinct components: a latent Planner and a Decoder for verbalization.
The Planner autoregressively evolves a trajectory of continuous states, while the Decoder provides text readouts when needed.
The Decoder grounds these latent plans onto the language space via a reconstruction objective.
This enables answer-aware adaptive termination and makes intermediate planning states available for textual readout, with a problem-dependent number of latent groups.
Our empirical evaluations on mathematical benchmarks reveal a distinctive behavioral pattern. 
We identify a trade-off between single-sample precision and sampled coverage: PLaT-8 has lower Pass@1 but competitive or stronger Pass@$k$ at larger $k$ relative to the compared latent methods in several settings. Pass@$k$ measures whether a correct candidate is present in a sampled pool, making it relevant to subsequent verification, reranking, voting, or search. Realizing this coverage as final-answer accuracy requires effective candidate selection.
Furthermore, the latent states in PLaT can be decoded into text for intermediate inspection without disrupting the continuous reasoning flow.
Our main contributions are summarized as follows:

\textbullet \enspace We formulate latent reasoning as an autoregressive sequence of continuous planning states with a textual Decoder that has separate parameters and is jointly optimized with the Planner, keeping generated text outside the Planner recurrence.

\textbullet \enspace We enable problem-dependent latent chain lengths through answer-aware stopping, while retaining textual access to intermediate planning states throughout the rollout.

\textbullet \enspace Experiments reveal a precision--coverage trade-off: PLaT can provide useful candidate coverage at larger $k$ in several settings, accompanied by lower Pass@1, motivating future verifier- and search-based inference.

\section{Related Work}

\subsection{Chain-of-thought Reasoning}
Chain-of-Thought (CoT) prompting enables large language models to solve complex problems by explicitly decomposing them into intermediate reasoning steps \cite{wei2022chain}.
Subsequent work shows that CoT can be activated with minimal prompting by a single instruction such as ``Let’s think step by step'' \cite{kojima2022large}.
Beyond simple CoT strategies, self-consistency \cite{wang2023selfconsistency} samples multiple CoT trajectories and marginalizes over final answers, improving robustness by approximating a distribution over reasoning paths.
Tree-of-Thought (ToT) \cite{yao2023tree,long2023large,mo2024tree} generalizes CoT into an explicit tree search, allowing the model to branch, evaluate, and backtrack over intermediate states.
Graph-of-Thought (GoT) \cite{yao2023beyond,besta2024graph,yao2024got} further extends this paradigm to graph-structured reasoning, enabling the reuse and recombination of intermediate conclusions across different branches.
Orthogonal to structured search, other works focus on refining intermediate reasoning steps through iterative revision or verification.
These approaches improve correctness by critiquing, editing, or validating generated reasoning traces, often in a multi-pass manner \cite{madaan2023self,lyu2023faithful,yang2023large}.

\subsection{Latent Reasoning}
Recent latent reasoning methods aim to reduce the cost and path collapse of explicit CoT by shifting part of the reasoning process into continuous hidden states, while retaining the final answer in natural language.
Early approaches explore partial internalization of reasoning, where additional non-semantic structures are introduced during inference.
For example, pause tokens or planning tokens allow models to suspend surface-level generation and perform internal processing before producing the next reasoning step \cite{goyal2024think,wang2024guiding}.
A second line of work focuses on compressing explicit CoT into latent representations, progressively removing or softening intermediate textual steps.
Curriculum-based approaches such as Coconut gradually replace explicit reasoning tokens with continuous latent states, enabling the model to internalize multi-step reasoning \cite{hao2025training}.
Related techniques distill explicit CoT trajectories into latent spaces or employ continuous vectors for intermediate reasoning steps \cite{zhang2025soft,shen2025codi}.
Other approaches further explore latent compression objectives to stabilize and optimize implicit reasoning processes \cite{tan2025think}.
These approaches seek to reduce explicit reasoning generation, typically treating latent states as end-to-end optimized carriers of reasoning with limited interfaces for inspecting intermediate states.

\section{Method}
\label{sec:method}
In this section, we first formalize reasoning as a latent autoregressive process. 
We then detail the PLaT architecture, the training via reconstruction, and inference and termination.
All notations are listed in Appendix \ref{sec:appendix-notation}.

\begin{figure}[htbp]
    \centering
    \includegraphics[width=\linewidth]{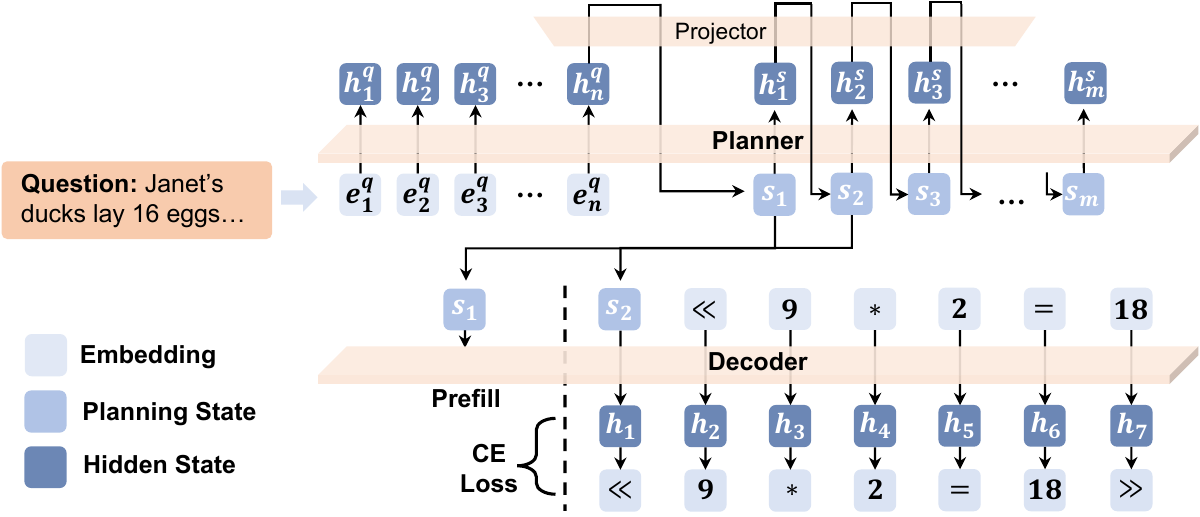}
    \caption{\textbf{Framework of the proposed PLaT paradigm.} The Planner autoregressively generates latent states conditioned on the question, and an independent Decoder uses each latent group as a soft prefix to generate a text segment. The illustration shows a general $L=2$ example, with two latent vectors per decoded segment; the main experimental configuration uses $L=1$.}
    \label{fig:method-framework}
\end{figure}

\subsection{Problem Formulation: Reasoning as Latent Planning}
Standard CoT models the probability of a reasoning chain $y = (y_1, \dots, y_T)$ autoregressively in the discrete token space:
\begin{equation}
    p(y|x) = \prod_{k=1}^T p(y_k | y_{<k}, x)
\end{equation}
Sampling from this formulation commits each trajectory to a specific token at each step, although other continuations remain available across samples.

PLaT introduces a sequence of continuous latent variables corresponding to textual segments.
We tokenize the reasoning text and divide it into $N$ segments of at most $K$ tokens; segment boundaries need not coincide with complete equations.
Let $\mathbf{S}_k=(\mathbf{s}_{k,1},\dots,\mathbf{s}_{k,L})$ denote the $L$ latent vectors assigned to segment $y_k$.
The Planner deterministically generates these vectors from the question and preceding latent vectors, without feeding decoded text back into the Planner. The latent history is maintained autoregressively by the Planner.
For a given segmentation into $N$ groups, the Decoder readout likelihood factorizes as follows:
\begin{equation}
\label{eq:joint_prob}
p(y|x)=\prod_{k=1}^{N}p_{\mathcal{D}}(y_k|\mathbf{S}_k(x)).
\end{equation}
The main configuration uses $K=8$ and $L=1$; the number of latent groups $N$ is determined dynamically during inference.

\subsection{Architecture}
\label{sec:architecture}
As shown in Figure~\ref{fig:method-framework}, PLaT implements the above formulation through two distinct modules: the Planner and the Decoder.
The Planner $\mathcal{M}$ and independent Decoder $\mathcal{D}$ have separate parameters.
A latent-generation projection $\phi$ maps Planner hidden states back to the backbone embedding dimension $d_m$.
It consists of a linear layer to width $d_p$, GELU, a linear layer back to $d_m$, and LayerNorm; projection dropout is zero in the verified configurations.

\paragraph{Planner.}
The Planner is responsible for evolving the reasoning trajectory autoregressively on the latent manifold.
The final hidden state of the question initializes the first latent vector, without appending an additional start token:
\begin{equation}
\mathbf{s}_{1,1}=\phi\!\left(\mathcal{M}(x)_{-1}\right).
\end{equation}
Each subsequent latent vector is generated by feeding the previous vector as an input embedding while retaining the question and latent-history key-value cache:
\begin{equation}
\mathbf{s}_{k,i}=\phi\!\left(\mathcal{M}([x;\mathbf{S}_{<k};\mathbf{s}_{k,<i}])_{-1}\right),
\qquad (k,i)\ne(1,1).
\end{equation}
It is important to note that the Planner generates deterministic vectors for a fixed question. Sampling diversity arises at the Decoder: each latent group defines a distribution over possible textual continuations.

\paragraph{Decoder.}
The Decoder maps each latent group to a distribution over textual segments.
It takes the current latent group $\mathbf{S}_k$ directly as a soft prefix for generating the text segment $y_k$:
\begin{equation}
P(y_k|\mathbf{S}_k)=\prod_{j=1}^{|y_k|}P_{\mathcal{D}}(y_{k,j}|\mathbf{S}_k,y_{k,<j}).
\end{equation}
The main configuration uses no EMA aggregation.
The Decoder conditions only on the current latent group and its own within-segment tokens; the question and earlier decoded segments are not supplied as additional context.
Decoder-context variants are evaluated separately in the ablation study.

\subsection{Supervised Training via Reconstruction}
During Supervised Fine-Tuning (SFT), we optimize the entire pipeline end-to-end using a latent-to-text reconstruction loss.
The loss is calculated as the cross-entropy between the ground-truth text $y_k$ and the model's prediction conditioned on the state $\mathbf{S}_k$:
\begin{equation}
    \mathcal{L}_{\text{rec}} = - \sum_{k=1}^{N} \sum_{j=1}^{|y_k|} \log P(y_{k,j} | \mathbf{S}_k, y_{k,<j}).
\end{equation}
This formulation treats intermediate reasoning steps and the final answer uniformly within the latent space, eliminating the need for mode-switching mechanisms and the constraint of a fixed number of latent steps.
No latent-history noise is injected in the main configuration. The Planner LoRA parameters, latent projection, and independent Decoder are optimized through the reconstruction objective.

\subsection{Inference and Termination}
Since the Planner operates in the latent space, its trajectory can advance without feeding generated text back into the Planner.
Each latent group is decoded for up to $K$ tokens; $K$ controls the local text budget rather than the total number of latent groups.
We detect the tokenized prefix \texttt{The answer is:} in the accumulated output, allowing the prefix to span segment boundaries.
After that prefix is detected, generation stops when EOS occurs in the answer content, or when the configured rollout limit is reached. This is an answer-aware rollout mechanism rather than a learned latent halting head.
The number of latent groups is therefore not fixed by $K$.

\section{Experiments}

\subsection{Experimental Setup.}
\paragraph{Model.}
For the GSM8K-trained comparison, we use GPT-2 \cite{radford2019language} and the instruction-tuned Llama-3.2-1B and Llama-3.2-3B models \cite{meta2024llama32}. For the separate OpenR1-Math setting, all methods use only Llama-3.2-1B-Instruct. This provides a focused comparison of longer-form reasoning at a fixed backbone size while keeping long-sequence training and 128-sample evaluation computationally manageable. The OpenR1 experiments therefore assess performance on longer-form tasks at the 1B scale; the GSM8K experiments provide the comparison across backbone sizes.
\paragraph{Datasets.}
We train on \textbf{GSM8K-Aug} \cite{deng2023implicit}, which augments \textbf{GSM8K} \cite{cobbe2021gsm8k} with equation-formatted reasoning generated by GPT-4 \cite{achiam2023gpt}. GSM8K contains multi-step grade-school mathematics problems; we evaluate on its 1,319 test questions. To assess transfer beyond the training distribution, we also use \textbf{GSM-Hard} \cite{gao2022pal}, which increases numerical difficulty, \textbf{SVAMP} \cite{patel-etal-2021-nlp}, which tests robustness to variations in arithmetic word problems, and \textbf{MultiArith} \cite{roy2015solving}, which requires combining multiple arithmetic operations.

For longer-form reasoning, we train separately on \textbf{OpenR1-Math-220k} \cite{openr1math}, whose extended mathematical solutions provide a less regularly segmented form of supervision. We evaluate on \textbf{MATH-500} \cite{hendrycks2021measuring}, comprising 500 competition-level problems, and \textbf{AIME 2024} and \textbf{AIME 2025} \cite{aime_problems_solutions}, with 30 competition problems each. This setting tests whether the approach extends beyond short equation-formatted solutions and is reported separately from the GSM8K-trained comparison. Data preparation and grading details appear in Appendix~\ref{sec:data-eval-details}.

\paragraph{Baselines.}
We benchmark PLaT against three representative baselines:
(1) \textbf{CoT-SFT} \cite{wei2022chain}: Fine-tunes the model with reasoning chains, and the trained models generate step-by-step during inference.
(2) \textbf{Coconut} \cite{hao2025training}: A curriculum-based latent reasoning method that progressively replaces explicit tokens with implicit hidden states.
(3) \textbf{CODI} \cite{shen2025codi}: A latent reasoning framework that employs hidden-state distillation from an explicit reasoning teacher.
All tables identify the backbone and use the corresponding fixed trained model.
The Llama-1B PLaT row uses the PLaT-8 configuration.
Coconut Llama-1B uses a stabilized learning rate of $3\times10^{-5}$ rather than the original $10^{-4}$ setting.
CODI Llama-3B uses the best available model supported by validation results; the optimum over the complete training history could not be verified.

\begin{table}[t]
\centering
\scriptsize
\setlength{\tabcolsep}{1.5pt}
\renewcommand{\arraystretch}{1.15}
\caption{\textbf{Main results on the GSM8K-trained models (\%).} Pass@$k$ is computed from the first $k$ of 128 samples within each run, then averaged over the three runs. Standard deviations are given in Appendix~\ref{sec:full-results}. Gray CoT-SFT rows are excluded from the latent-method ranking. Bold and underlined entries mark the highest and second-highest displayed latent-method scores per column, including ties. PLaT is evaluated with $K=8$ (PLaT-8).}
\label{tab:exp-main}
\resizebox{\linewidth}{!}{%
\begin{tabular}{llrrrr|rrrr|rrrr|rrrr}
\toprule
\multirow{2}{*}{Model} & \multirow{2}{*}{Method} & \multicolumn{4}{c}{GSM8K} & \multicolumn{4}{c}{GSM-Hard} & \multicolumn{4}{c}{SVAMP} & \multicolumn{4}{c}{MultiArith} \\
\cmidrule(lr){3-6}\cmidrule(lr){7-10}\cmidrule(lr){11-14}\cmidrule(lr){15-18}
 & & @1 & @32 & @64 & @128 & @1 & @32 & @64 & @128 & @1 & @32 & @64 & @128 & @1 & @32 & @64 & @128 \\
\midrule
\rowcolor{gray!15} \multirow{4}{*}{GPT-2} & CoT-SFT & 35.30 & 74.20 & 80.41 & 86.33 & 7.10 & 18.35 & 20.65 & 22.52 & 36.63 & 65.37 & 70.50 & 75.47 & 85.00 & 99.07 & 99.44 & 99.63 \\
 & Coconut & \underline{32.94} & 55.69 & 61.45 & 66.19 & \underline{7.62} & 13.65 & 15.01 & 16.15 & \underline{35.25} & 50.15 & 53.05 & 56.45 & \underline{81.67} & \underline{90.83} & \underline{91.94} & \underline{93.33} \\
 & CODI & \textbf{40.16} & \textbf{63.00} & \underline{67.12} & \underline{69.90} & \textbf{8.74} & \textbf{14.58} & \underline{15.52} & \underline{16.53} & \textbf{38.60} & \underline{53.67} & \underline{55.80} & \underline{58.13} & \textbf{92.04} & \textbf{96.67} & \textbf{96.67} & \textbf{96.67} \\
 & PLaT & 18.35 & \underline{59.51} & \textbf{67.74} & \textbf{76.08} & 3.75 & \underline{14.22} & \textbf{17.06} & \textbf{18.88} & 21.80 & \textbf{53.70} & \textbf{60.75} & \textbf{66.45} & 33.61 & 79.44 & 87.50 & 91.94 \\
\midrule
\rowcolor{gray!15} \multirow{4}{*}{Llama-1B} & CoT-SFT & 49.96 & 86.62 & 90.45 & 93.18 & 11.45 & 23.68 & 25.50 & 27.72 & 59.50 & 87.97 & 91.07 & 93.37 & 94.07 & 100.00 & 100.00 & 100.00 \\
 & Coconut & 27.52 & 49.28 & 53.68 & 58.07 & 6.47 & 11.95 & 13.17 & 14.18 & 41.87 & 53.13 & 54.90 & 56.97 & 59.07 & 85.93 & 87.41 & 89.44 \\
 & CODI & \textbf{54.55} & \textbf{69.11} & \underline{71.08} & \underline{72.37} & \textbf{12.24} & \textbf{16.41} & \underline{17.13} & \underline{17.55} & \textbf{58.75} & \textbf{70.70} & \textbf{72.15} & \underline{73.00} & \textbf{97.22} & \textbf{98.33} & \textbf{98.33} & \underline{98.33} \\
 & PLaT & \underline{36.20} & \underline{68.73} & \textbf{75.28} & \textbf{80.59} & \underline{8.26} & \underline{16.30} & \textbf{18.23} & \textbf{19.75} & \underline{43.00} & \underline{66.87} & \underline{70.63} & \textbf{74.47} & \underline{84.63} & \underline{96.85} & \underline{97.41} & \textbf{98.70} \\
\midrule
\rowcolor{gray!15} \multirow{4}{*}{Llama-3B} & CoT-SFT & 58.88 & 92.90 & 95.27 & 96.71 & 18.95 & 38.34 & 41.37 & 44.12 & 74.67 & 95.50 & 96.90 & 97.83 & 97.78 & 100.00 & 100.00 & 100.00 \\
 & Coconut & 28.38 & 62.32 & 70.18 & 75.87 & 6.09 & 15.62 & 17.29 & 19.13 & 34.07 & 57.47 & 61.73 & 65.43 & 89.26 & 97.04 & 97.22 & 97.96 \\
 & CODI & \textbf{57.16} & \textbf{76.72} & \textbf{80.44} & \underline{83.24} & \textbf{13.38} & \textbf{18.12} & \textbf{19.45} & \textbf{20.89} & \textbf{68.70} & \textbf{80.65} & \textbf{82.60} & \textbf{84.60} & \textbf{99.72} & \textbf{100.00} & \textbf{100.00} & \textbf{100.00} \\
 & PLaT & \underline{39.85} & \underline{73.82} & \underline{80.09} & \textbf{84.79} & \underline{9.48} & \underline{17.66} & \underline{19.23} & \underline{20.87} & \underline{54.77} & \underline{76.00} & \underline{79.43} & \underline{82.87} & \underline{95.37} & \underline{98.89} & \underline{99.63} & \underline{99.63} \\
\bottomrule
\end{tabular}%
}
\end{table}

\begin{table}[!b]
    \centering
    \caption{\textbf{Efficiency comparison.} The PLaT column reports lazy, greedy inference with the GPT-2 backbone on 200 GSM8K test questions. Forward counts include all backbone calls during planning and readout; times are per-question means $\pm$ across-question standard deviations. All methods are measured on the same hardware.}
    \label{tab:exp-time}
    \small
    \begin{tabular}{l|rrrr}
    \toprule
        & \textbf{CoT} & \textbf{Coconut} & \textbf{CODI} & \textbf{PLaT} \\ \midrule
        \textbf{Fwd.} & 25.55 & 6.00 & 6.00 & 12.83 \\
        \textbf{Time (ms)} & $349.6_{\pm 8.9}$ & $100.6_{\pm 3.2}$ & $240.0_{\pm 17.2}$ & $237.5_{\pm59.2}$ \\ \bottomrule
    \end{tabular}
\end{table}

\paragraph{Evaluation.}
We report \textbf{Pass@$k$} ($k=1,32,64,128$), the fraction of questions with at least one correct answer among the first $k$ of 128 sampled responses. Short-form evaluations use temperature $0.9$ and a 256-token generation cap. We average over available sampling runs and report their standard deviations in Appendix~\ref{sec:full-results}; full evaluation settings appear in Appendix~\ref{sec:data-eval-details}. Pass@$k$ measures candidate coverage; realizing this coverage as final-answer accuracy requires answer selection.

\paragraph{Implementation Details.}
Unless otherwise stated, PLaT training uses a learning rate of $8\times10^{-4}$ with BF16 computation, LoRA adaptation \cite{hu2022lora} (rank 128, $\alpha=32$) on the Planner, and a full-parameter Decoder with separate parameters, jointly optimized with the Planner through the reconstruction objective. We denote the variant with $K$ tokens per segment by PLaT-$K$; the suffix specifies segment length, not the number of latent vectors $L$ or the total reasoning length. The main configuration, PLaT-8, uses one latent vector per eight-token segment ($L=1$). Additional settings are provided in Appendix~\ref{sec:appendix-exp-detail}.

\subsection{Effectiveness}

\paragraph{Main results.}
Among the compared latent methods, PLaT's main advantage is broader candidate coverage under repeated sampling, accompanied by lower single-sample accuracy (Table~\ref{tab:exp-main}). On GSM8K, PLaT-8 reaches 80.59\% Pass@128 with Llama-1B versus CODI's 72.37\%, despite lower Pass@1 (36.20\% versus 54.55\%). The Llama-3B comparison shows the same trade-off, with a smaller coverage advantage. This pattern makes PLaT relevant when a system can examine several candidates, but does not establish an advantage when only one answer is generated. The benefit is also task-dependent: on Llama-3B SVAMP, CODI retains higher coverage (84.60\% versus 82.87\%). CoT-SFT achieves higher Pass@1 and Pass@128 than PLaT across all backbones and datasets in Table~\ref{tab:exp-main}; the coverage gains therefore establish an advantage over particular latent baselines, not over explicit CoT.

\paragraph{Sampling behavior.}
The gap between Pass@1 and Pass@$k$ indicates that correct answers often appear outside the first sampled response. In PLaT, these alternatives arise from repeated Decoder readouts of a deterministic latent trajectory. The results therefore motivate using this readout distribution to supply candidates for verification or search. The gains from additional samples make this distinction concrete. On Llama-1B GSM8K, increasing the pool from 32 to 128 responses raises PLaT's coverage from 68.73\% to 80.59\%, a gain of 11.86 percentage points, compared with 3.26 points for CODI (69.11\% to 72.37\%). PLaT is slightly behind at 32 samples and ahead at 64. Thus, within this sampling range, additional readouts continue to solve questions missed by the smaller pool. This is a benefit in candidate availability; converting it into final-answer accuracy still requires effective selection. The OOD results further indicate that the usefulness of this candidate pool depends on the task and backbone. Per-cell uncertainty is reported in Appendix~\ref{sec:full-results}.

\subsection{Efficiency}

Table~\ref{tab:exp-time} shows that selective verbalization reduces GPT-2 latency relative to CoT on the same hardware: 237.5 versus 349.6 ms per question, a 32.1\% reduction. Lazy readout skips intermediate text after a failed answer-prefix probe, while retaining Decoder calls for answer detection and generation. The textual interface thus remains available without fully verbalizing each intermediate state.

PLaT is comparable in measured latency to CODI (240.0 ms), but slower than Coconut (100.6 ms). Its 12.83 backbone calls fall between CoT's 25.55 and the latent baselines' 6.00, illustrating that call counts alone do not determine latency across architectures. These greedy measurements characterize selective readout cost; multi-sample efficiency remains to be evaluated.

\subsection{Analysis of Latent States}

\paragraph{Evolution of Reasoning Diversity.}
A deterministic planning state can support several distinct textual readouts. On all 1,319 GSM8K test questions, we sample 10 readouts per trajectory position and compare their diversity with explicit CoT. PLaT averages 4.648 distinct chunks and 4.347 reference-compatible chunks per nonempty state group. The proximity of these means suggests that much of the observed textual diversity remains compatible with the reference reasoning under the judgment protocol, rather than consisting only of invalid fragments.

\begin{figure}[tbp]
\centering
\begin{subfigure}[t]{0.495\linewidth}
    \centering
    \includegraphics[width=\linewidth]{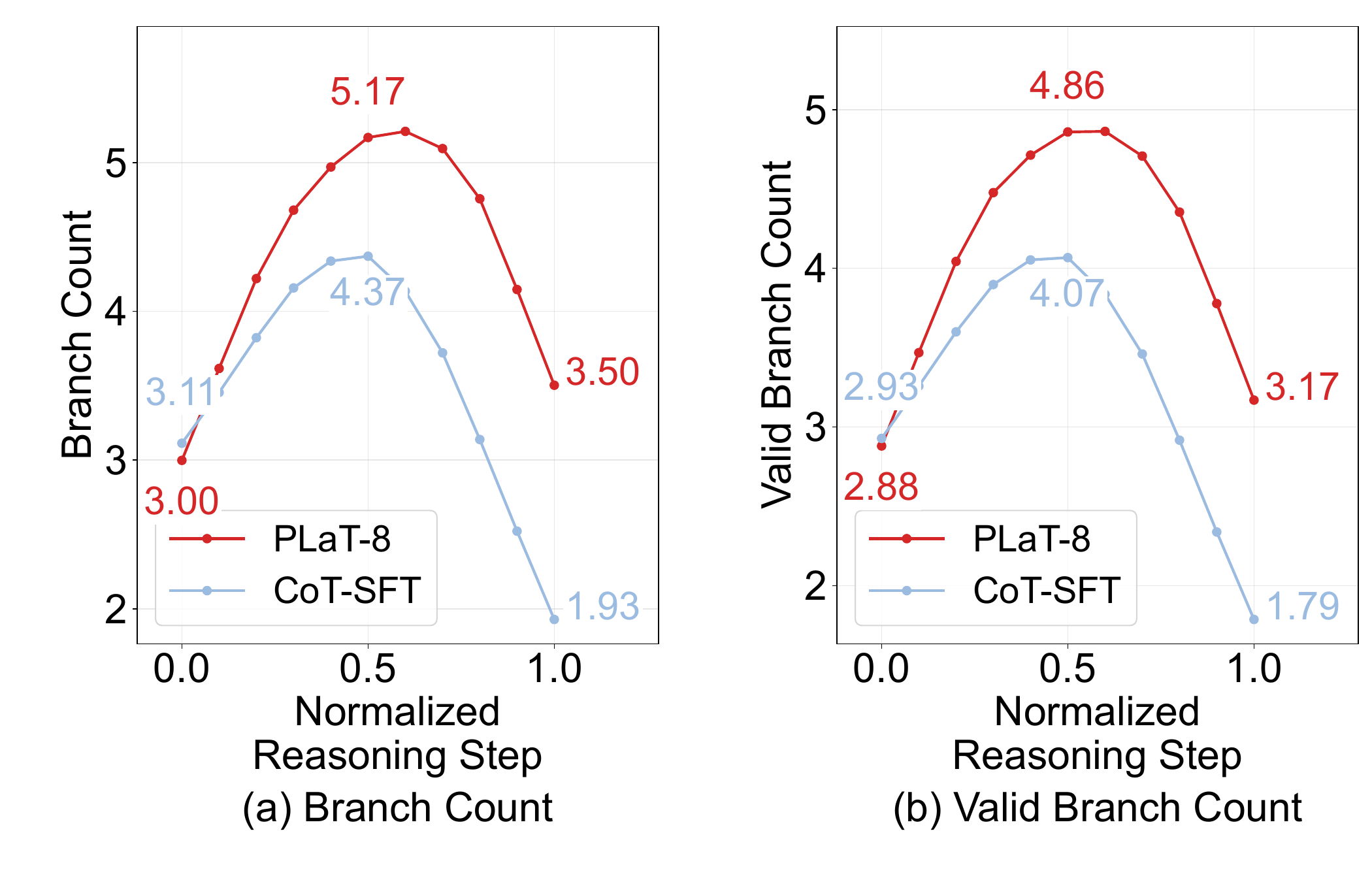}
    \caption{Evolution of exploration during reasoning.}
    \label{fig:exp-branch-a}
\end{subfigure}\hfill
\begin{subfigure}[t]{0.495\linewidth}
    \centering
    \includegraphics[width=\linewidth]{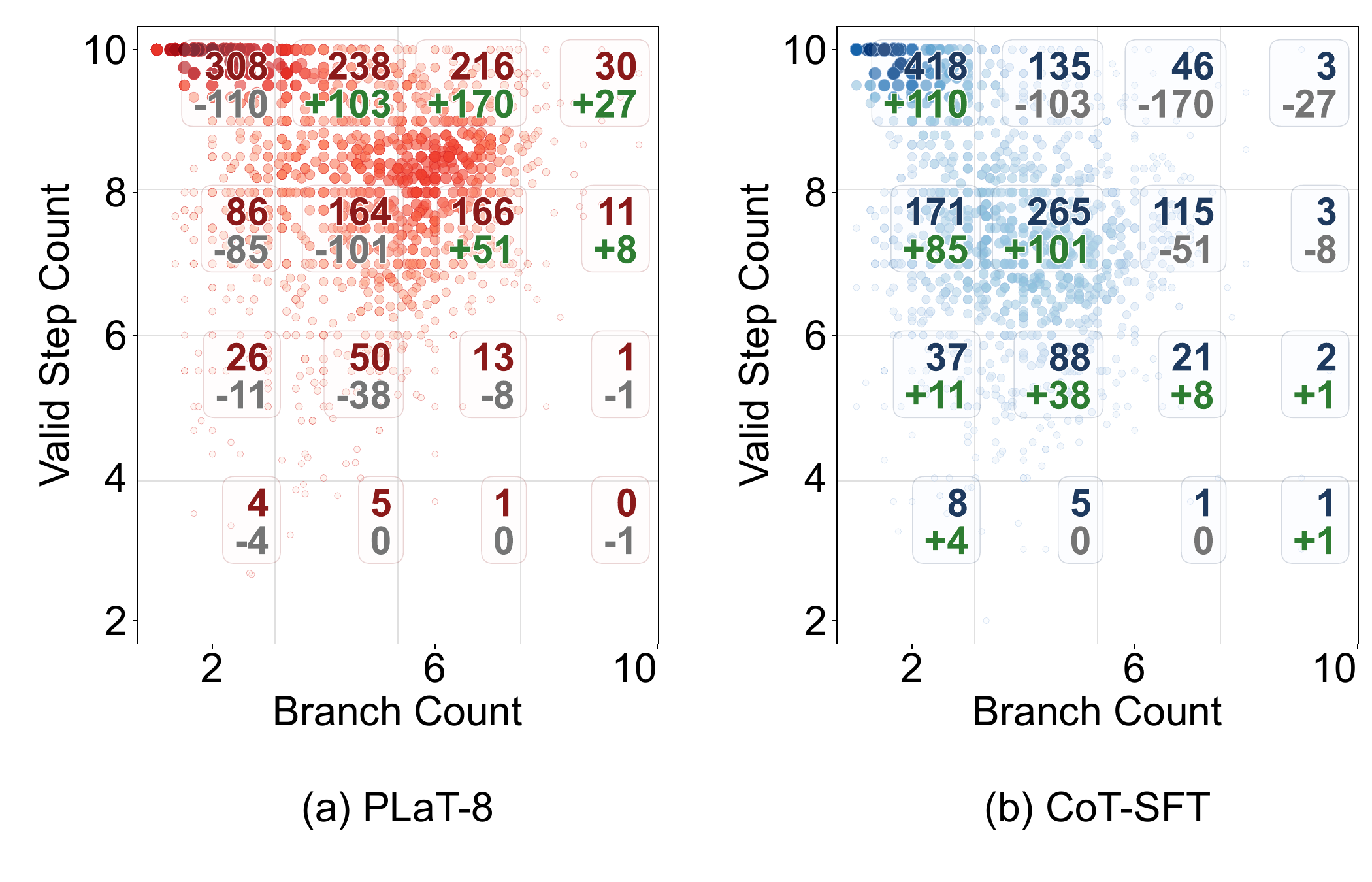}
    \caption{Branching factor versus valid step count.}
    \label{fig:exp-branch-b}
\end{subfigure}
\caption{\textbf{Textual readout branching from latent states.} Results use PLaT and CoT-SFT on 1,319 questions with 10 sampled paths per question. For PLaT, the Planner trajectory is fixed for a question; the reported branches are distinct textual continuations sampled from intermediate latent states. In (a), curves show distinct and valid readout counts after normalizing each question's state sequence to 11 positions. In (b), samples without intermediate steps are excluded; the 16 zones annotate counts and between-method differences.}
\label{fig:exp-branching}
\end{figure}

Figure~\ref{fig:exp-branching} shows that PLaT's additional readout diversity emerges as reasoning progresses. Both methods begin with similar counts, but PLaT retains more alternatives through the middle and later positions. At the final normalized position, PLaT still exposes 3.50 distinct readouts versus 1.93 for CoT-SFT; the corresponding valid-readout counts are 3.17 and 1.79. Thus, the separation persists after validity filtering and extends to the end of the trajectory. The scatter plots complement this aggregate trend: PLaT has more questions in regions combining higher branch counts with high valid-step counts. Together, these patterns suggest that a fixed Planner trajectory can sustain multiple locally compatible textual continuations across successive positions. This provides a readout-level perspective on the candidate diversity observed in the Pass@$k$ results, without establishing distinct solution strategies or a causal explanation of the coverage gains. Counting and judgment details appear in Appendix~\ref{sec:readout-count-details}.

\paragraph{Distribution of Exploration vs. Exploitation.}
Figure~\ref{fig:exp-branch-b} shows how readout diversity and local validity coexist across questions. CoT-SFT concentrates more questions in the upper-left region, where many sampled chunks are valid but yield relatively few distinct continuations. PLaT shifts more of this mass toward the upper-right: the two rightmost cells in the highest-validity row contain 246 questions for PLaT, compared with 49 for CoT-SFT. Conversely, the leftmost cell in that row contains 308 PLaT questions versus 418 for CoT-SFT. The contrast shows that high local validity can accompany a broader range of textual readouts from PLaT's fixed planning trajectory. The additional diversity is therefore visible across questions with consistently valid local readouts, supporting the use of the Decoder as an interface for generating alternative continuations. These are local textual alternatives, not evidence that each continuation forms a distinct, correct complete solution. The judgment protocol and human-review results are provided in Appendices~\ref{sec:readout-count-details} and~\ref{sec:appendix-agree-judge}.

\subsection{Ablation Study}
\begin{table}[tb]
\centering
\small
\setlength{\tabcolsep}{3pt}
\caption{\textbf{PLaT architectural and segmentation variants on Llama-1B GSM8K (\%).} Entries report mean $\pm$ sample standard deviation over multiple verified sampling runs. Decoder-context variants use PLaT-8 with $L=1$; fixed-$K$ and latent-count sweeps are reported separately.}
\begin{tabular}{lrrrr}
\toprule
& Pass@1 & Pass@32 & Pass@64 & Pass@128 \\
\midrule
PLaT-8 & $36.20\pm0.38$ & $68.73\pm0.05$ & $75.28\pm0.00$ & $80.59\pm0.32$ \\
PLaT semantic-step & $37.86\pm1.10$ & $69.45\pm0.46$ & $75.16\pm0.62$ & $80.82\pm0.62$ \\
CoT initialization & $29.95\pm0.43$ & $67.48\pm0.86$ & $74.15\pm0.43$ & $78.01\pm0.11$ \\
Decoder latent history & $38.51\pm0.11$ & $67.29\pm0.70$ & $73.62\pm0.21$ & $78.85\pm0.32$ \\
Decoder question + latent history & $31.61\pm0.95$ & $67.73\pm0.99$ & $74.37\pm0.23$ & $80.29\pm0.82$ \\
\bottomrule
\end{tabular}
\label{tab:exp-ablation}
\end{table}
Table~\ref{tab:exp-ablation} supports fixed-token segmentation as a practical alternative to semantic-step supervision: their Pass@128 values are close on GSM8K, although the training differences preclude attributing the comparison solely to segmentation. Our choice of fixed-token groups is motivated by portability. Equation-formatted GSM8K provides clear step boundaries, whereas longer solutions such as OpenR1-Math do not provide uniform step annotations. Controlling segment length through $K$ allows the same interface to extend to such data without a task-specific semantic splitter.

Additional Decoder context changes the precision--coverage trade-off rather than improving both metrics consistently. Latent history raises Pass@1 but lowers Pass@128; adding the question as well does not improve either metric over PLaT-8. Both variants preserve decoupling because generated text remains outside the Planner recurrence. These results therefore do not establish that local conditioning alone explains PLaT's lower Pass@1. The CoT-initialized variant also falls below PLaT-8, but other training differences prevent isolating initialization as the cause. Full results across datasets appear in Appendix~\ref{sec:ablation-single}.

\subsection{Analysis of Hyperparameters}
We analyze token-group size $K$ and latent vectors per segment $L$ separately from the architectural ablations. Here, $K$ sets the text span assigned to each latent group, while $L$ sets the number of vectors representing that span. Neither fixes the total number of groups $N$, which can vary across questions under answer-aware stopping. These sweeps therefore examine local representation granularity, separately from the ability to extend the overall latent chain.
The four-dataset tables in Appendix~\ref{sec:ablation-single} aggregate three verified sampling runs per configuration and report sample standard deviations. The sensitivity plots retain their original single-run visualizations and are intended to show trends rather than establish monotonic effects.

\paragraph{Token-group size ($K$).}
With one latent vector per segment, we compare $K\in\{2,5,8,16,32\}$. Across datasets, the intermediate settings $K=5$ or $K=8$ generally retain more sampled coverage than the extremes; the results are shown in Appendix Fig.~\ref{appendix_fig:hyper-token-size}.

\paragraph{Latent vectors per segment ($L$).}
Increasing the number of latent vectors per segment does not consistently improve coverage. At $K=8$, the $L\in\{1,2,3\}$ sweep yields dataset-dependent changes that are smaller than those from changing $K$ on several tasks (Appendix Fig.~\ref{appendix_fig:hyper-latent-num}). These results do not support treating additional latent vectors as a uniformly beneficial increase in capacity.

\subsection{Longer-form Mathematical Evaluation}
\begin{table}[ht]
\centering
\caption{OpenR1-Math-trained Llama-1B results on longer-form benchmarks (Pass@1 and Pass@128, \%). Each dataset reports the two prefix-coverage metrics separately. All entries report means $\pm$ sample standard deviations over three complete sampling runs. Gray CoT-SFT is excluded from ranking; bold and underlined scores denote the best and second-best latent-method results in each column, with ties sharing a mark. This table is a separate evaluation setting and is not directly comparable with the GSM8K-trained main table.}
\resizebox{\linewidth}{!}{%
\begin{tabular}{lrrrrrr}
\toprule
Method & \multicolumn{2}{c}{MATH-500} & \multicolumn{2}{c}{AIME 2024} & \multicolumn{2}{c}{AIME 2025} \\
 & Pass@1 & Pass@128 & Pass@1 & Pass@128 & Pass@1 & Pass@128 \\
\midrule
\rowcolor{gray!15} CoT-SFT & $16.20\pm0.28$ & $77.70\pm0.99$ & $1.67\pm2.36$ & $15.00\pm2.36$ & $1.67\pm2.36$ & $20.00\pm4.71$ \\
Coconut & $\underline{5.20}\pm0.85$ & $27.60\pm0.57$ & $\underline{0.00}\pm0.00$ & $3.33\pm0.00$ & $\underline{0.00}\pm0.00$ & $\underline{10.00}\pm0.00$ \\
CODI & $\textbf{7.70}\pm0.71$ & $\underline{38.70}\pm0.71$ & $\textbf{1.67}\pm2.36$ & $\textbf{18.33}\pm2.36$ & $\underline{0.00}\pm0.00$ & $3.33\pm4.71$ \\
\midrule
PLaT-16 & $2.50\pm0.14$ & $\textbf{44.80}\pm2.55$ & $\underline{0.00}\pm0.00$ & $\underline{11.67}\pm7.07$ & $\textbf{1.67}\pm2.36$ & $\textbf{16.67}\pm0.00$ \\
PLaT-32 & $0.80\pm0.00$ & $35.60\pm1.13$ & $\underline{0.00}\pm0.00$ & $6.67\pm0.00$ & $\underline{0.00}\pm0.00$ & $8.33\pm7.07$ \\
\bottomrule
\end{tabular}
}
\end{table}

The longer-form results retain the coverage--precision trade-off but show its limits. On MATH-500, PLaT-16 has the highest Pass@128 among the evaluated latent methods (44.80\%), while its Pass@1 remains below CODI and Coconut. CoT-SFT is substantially stronger on both metrics, so this comparison does not establish parity with explicit reasoning. Increasing segment length to PLaT-32 lowers coverage on all three benchmarks. The AIME results are mixed and based on only 30 questions per year, providing limited evidence for broader generalization to competition mathematics.

\section{Conclusion}
PLaT separates the autoregressive evolution of continuous planning states from textual readout through an independent Decoder. Generated text remains outside the Planner recurrence, and answer-aware stopping allows the number of latent groups to vary without fixing the total chain length through the local segment configuration.

The experiments identify a precision--coverage trade-off: PLaT provides stronger candidate coverage at larger $k$ than the compared latent methods in several settings, while baseline methods often achieve higher Pass@1 and CoT-SFT remains stronger overall. This pattern motivates PLaT's potential role as a candidate generator, where additional readouts can expose correct answers missed by smaller sample pools.

The independent readout interface enables the study of planning-state content and textual diversity along a latent trajectory. Establishing causal faithfulness and translating candidate coverage into improved answer selection remain directions for further evaluation.

\subsection*{AI Use Statement}
Generative AI tools were used to assist with code development, figure preparation, and language polishing. All AI-assisted code, figures, and text were reviewed and verified by the authors, who take full responsibility for the final manuscript. 

\subsection*{Ethics Statement}
PLaT provides a textual interface for inspecting intermediate planning states, which may support the study of latent reasoning. The usefulness of this interface for auditing remains unverified: fluent readouts can be incorrect or fail to faithfully describe the computation that produced an answer, and should not be treated as verified explanations or guarantees of trustworthy decisions. High candidate coverage likewise requires reliable answer selection before it can improve deployed accuracy. The lazy-decoding timing result is limited to a shared-GPU greedy setting; broader end-to-end efficiency benefits remain to be established.

\subsection*{Reproducibility Statement}
Section~\ref{sec:method} specifies the latent recurrence, Decoder conditioning, reconstruction loss, and stopping mechanism. Appendix~\ref{sec:appendix-exp-detail} records implementation details, model selection and repeated-sampling result summaries. Appendices~\ref{sec:appendix-agree-judge} and~\ref{sec:appendix-prompt} describe the readout-judgment protocol and agreement analysis. The reported repetitions concern evaluation sampling rather than independent model training.

\bibliography{bibliography}
\bibliographystyle{iclr2027_conference}

\newpage

\appendix
\setcounter{figure}{0}
\setcounter{table}{0}

\section{Notations}
\label{sec:appendix-notation}
\begin{table}[ht]
    \centering
    \caption{Summary of notations used in PLaT.}
    \label{appendix_tab:notation}
    \begin{tabular}{l l l}
        \toprule
        \textbf{Symbol} & \textbf{Definition} & \textbf{Note} \\
        \midrule
        $x$ & Input question sequence & Token sequence \\
        $y$ & Complete reasoning chain & $N$ segments, $T$ tokens \\
        $y_k$ & The $k$-th textual segment & At most $K$ tokens \\
        $k$ & Index of textual segments & $k\in\{1,\dots,N\}$ \\
        $i$ & Index within a latent group & $i\in\{1,\dots,L\}$ \\
        $\mathcal{M}$ & Planner backbone & LoRA-adapted \\
        $\mathcal{D}$ & Independent Decoder & Fully trained \\
        $d_m$ & Backbone hidden dimension & Latent vectors have this dimension \\
        $d_p$ & Projection intermediate width & Between the two linear layers \\
        $T$ & Total number of text tokens & Before segmentation \\
        $N$ & Number of textual segments / latent groups & Variable across questions \\
        $K$ & Tokens per segment & Main configuration: 8 \\
        $L$ & Latent vectors per segment & Main configuration: 1 \\
        $\mathbf{h}$ & Backbone hidden state & $\mathbf{h}\in\mathbb{R}^{d_m}$ \\
        $\mathbf{s}_{k,i}$ & Latent vector & $\mathbf{s}_{k,i}\in\mathbb{R}^{d_m}$ \\
        $\mathbf{S}_k$ & Current Decoder prefix & $(\mathbf{s}_{k,1},\dots,\mathbf{s}_{k,L})$ \\
        $\phi$ & Latent-generation projection & $\mathbb{R}^{d_m}\to\mathbb{R}^{d_m}$ \\
        \bottomrule
    \end{tabular}
\end{table}

\section{Extra Information of Experiments}
\label{sec:appendix-exp-detail}
\subsection{More Implementation Details}
\paragraph{General Details.}
The GPT-2 and Llama-1B PLaT-8 configurations use LoRA dropout 0.05, one latent vector per eight-token segment, no state aggregation, and zero latent-history noise.
Neither configuration initializes the Planner or Decoder from CoT-SFT weights.
The projection dimension is 768 for GPT-2 and 2048 for Llama-1B, with projection dropout zero and LayerNorm enabled.
The training seed is 11. The fixed-token and semantic-step variants are trained and evaluated separately.
The sampling temperature for Pass@$k$ is set to $0.9$.
The stored evaluation settings use at most 97 latent rollout steps and 256 generated tokens, with full decoding for the reported PLaT-8 results.

\paragraph{Readout settings.}
The reported accuracy results use full decoding. For the efficiency measurement, an online lazy variant probes each latent group for the answer prefix, skips the remaining intermediate readout after a prefix mismatch, and resumes full fixed-token decoding once the answer prefix is detected. Probe calls are included in its computational cost. This is an approximate detection rule: a prefix beginning inside a skipped segment or spanning segments can be missed, so equivalence is evaluated empirically rather than assumed.

\subsection{Readout Counting Protocol}
\label{sec:readout-count-details}
We compared textual readout branching for PLaT against explicit CoT on the GSM8k test set (we did not use the OOD datasets because they do not contain intermediate-step annotations for reference).
For both methods, we sampled 10 reasoning paths per question (temperature=0.9) and analyzed them.
We study intermediate readouts from the fixed-token PLaT model and its CoT-SFT reference on all 1,319 GSM8K test questions, using 10 stored sampled trajectories per question.
The stored judgment pools contain 21,695 distinct PLaT chunks and 14,178 distinct CoT chunks.
The original 100-question trajectories and their judgments are retained; the remaining CoT questions come from a separate stored sampling batch.
GPT-5.4-mini receives the candidate and its stored previous/next readouts, together with reference reasoning.
This context is needed because an apparent equality can be the suffix of an expression split across chunk boundaries.
A truncated fragment is not invalid merely because it is incomplete.

The current pipeline distinguishes readout chunks by exact text, rather than semantic clustering.
Repeated chunk text is associated with its first stored context; this is not an occurrence-level judgment of every trajectory.
Consequently, these counts should be interpreted as diversity among sampled textual readouts, rather than as counts of independent Planner trajectories.
Judgments cover all distinct chunks in this analysis. Existing labels are preserved, and the first stored successful judgment is used for each newly judged text.
Mean branch counts are computed over 5,340 nonempty PLaT state groups and 5,979 nonempty CoT state groups, excluding empty readouts.
The full-test PLaT mean branch count is 4.648 and its mean valid branch count is 4.347; the previously reported 4.225 is retained as the mean valid branch count on the original 100-question subset.
The curves in Fig.~\ref{fig:exp-branching}(a) show distinct branch counts and distinct valid branch counts, normalize each question's state sequence to 11 positions, and then average across questions. Fig.~\ref{fig:exp-branching}(b) contains 1,319 questions; its valid-step count includes repeated sampled steps, unlike the distinct-branch count used in the curves.

\subsection{Data Preparation and Evaluation Details}
\label{sec:data-eval-details}
GSM8K-Aug supervision consists of a question, a sequence of equation-formatted reasoning steps, and an answer. PLaT divides the tokenized reasoning into segments of $K$ tokens, independently of semantic step boundaries; the main comparison uses $K=8$ and $L=1$.

We also evaluate on MATH-500, the 500-problem test split of MATH \cite{hendrycks2021measuring}, and the 30-problem AIME 2024 and AIME 2025 competition sets \cite{aime_problems_solutions}. For this longer-form setting, we train separate Llama-1B models on the OpenR1-Math-220k training release \cite{openr1math}. We use its cleaned \texttt{cot\_answer} field as reasoning supervision, preserve the independent answer field, and apply protected-boundary tokenization with a target block length of approximately 32 tokens and a preferred upper bound of 64 tokens for the latent variants. We evaluate on all 500 MATH-500 questions and all 30 questions from each AIME year, excluding these frozen test questions from training-data preparation. Each method uses one fixed trained model and 128 ordered samples per question; Pass@$k$ is the fraction of questions solved by at least one of the first $k$ samples. We use the recorded mathematical-equivalence grader (LaTeX2Sympy2 1.9.1 with ANTLR 4.7.2). These results are kept separate from the GSM8K-trained comparison because the training data, trained models, and sampling protocols differ.

Within a run, the $k$ samples form a candidate pool; they are not $k$ independent experimental repeats.
Each short-form evaluation run uses temperature $0.9$, 128 ordered samples per question, and a 256-token generation cap. PLaT uses top-$k=0$ and top-$p=1$. Table~\ref{tab:exp-main} reports means over the available sampling runs for each fixed trained model; Appendix~\ref{sec:full-results} reports sample standard deviations where multiple runs are available. These repetitions measure variation across evaluation runs, not variation from retraining. Method-specific prompts and evaluation batching are retained.

\subsection{Model Selection and Evaluation Protocol}
For each reported configuration, we evaluate a fixed set of trained weights. PLaT Llama-3B is selected by validation performance. The fixed-token and semantic-step variants are trained and evaluated separately; models are never substituted across configurations.

CODI GPT-2 uses the public model. CODI Llama-1B uses an adapted public model, evaluated with batch size 1 and disjoint sampling-seed ranges for its three repetitions. CODI Llama-3B uses the best available model supported by validation results; the optimum over the complete training history could not be verified.
Coconut Llama-1B uses learning rate $3\times10^{-5}$.
The comparisons retain method-specific prompts, batching, and runtime implementations.

\subsection{Sampling Repeats and Prefix Results}
\label{sec:full-results}
Each run uses one fixed trained model and 128 ordered samples per question. We first compute prefix Pass@$k$ within each run, then report the mean and sample standard deviation across sampling runs. These are sampling repeats, not independently trained models. Different evaluation runtimes are retained, so the standard deviations describe the available reruns rather than isolating seed effects.

\begin{table}[p]
\centering
\scriptsize
\setlength{\tabcolsep}{3pt}
\caption{GPT-2 prefix results (\%): mean $\pm$ sample standard deviation over repeated sampling runs.}
\resizebox{\linewidth}{!}{%
\begin{tabular}{llrrrr}
\toprule
Method & Dataset & Pass@1 & Pass@32 & Pass@64 & Pass@128 \\
\midrule
\rowcolor{gray!15} CoT-SFT & GSM8K & $35.30\pm0.54$ & $74.20\pm0.23$ & $80.41\pm0.29$ & $86.33\pm0.56$ \\
\rowcolor{gray!15} CoT-SFT & GSM-Hard & $7.10\pm0.44$ & $18.35\pm0.23$ & $20.65\pm0.49$ & $22.52\pm0.60$ \\
\rowcolor{gray!15} CoT-SFT & SVAMP & $36.63\pm0.93$ & $65.37\pm0.84$ & $70.50\pm0.82$ & $75.47\pm0.55$ \\
\rowcolor{gray!15} CoT-SFT & MultiArith & $85.00\pm0.00$ & $99.07\pm0.32$ & $99.44\pm0.00$ & $99.63\pm0.32$ \\
Coconut & GSM8K & $32.94\pm0.59$ & $55.69\pm0.16$ & $61.45\pm0.27$ & $66.19\pm0.64$ \\
Coconut & GSM-Hard & $7.62\pm0.05$ & $13.65\pm0.32$ & $15.01\pm0.11$ & $16.15\pm0.11$ \\
Coconut & SVAMP & $35.25\pm0.21$ & $50.15\pm0.78$ & $53.05\pm0.64$ & $56.45\pm0.49$ \\
Coconut & MultiArith & $81.67\pm0.79$ & $90.83\pm0.39$ & $91.94\pm1.18$ & $93.33\pm0.00$ \\
CODI & GSM8K & $40.16\pm1.18$ & $63.00\pm0.50$ & $67.12\pm0.19$ & $69.90\pm0.15$ \\
CODI & GSM-Hard & $8.74\pm0.19$ & $14.58\pm0.09$ & $15.52\pm0.12$ & $16.53\pm0.08$ \\
CODI & SVAMP & $38.60\pm0.95$ & $53.67\pm0.40$ & $55.80\pm0.36$ & $58.13\pm0.29$ \\
CODI & MultiArith & $92.04\pm0.85$ & $96.67\pm0.00$ & $96.67\pm0.00$ & $96.67\pm0.00$ \\
PLaT-8 & GSM8K & $18.35\pm0.11$ & $59.51\pm1.29$ & $67.74\pm0.80$ & $76.08\pm1.23$ \\
PLaT-8 & GSM-Hard & $3.75\pm0.05$ & $14.22\pm0.16$ & $17.06\pm0.64$ & $18.88\pm0.11$ \\
PLaT-8 & SVAMP & $21.80\pm0.71$ & $53.70\pm1.41$ & $60.75\pm1.91$ & $66.45\pm1.20$ \\
PLaT-8 & MultiArith & $33.61\pm2.75$ & $79.44\pm0.79$ & $87.50\pm0.39$ & $91.94\pm0.39$ \\
\bottomrule
\end{tabular}
}
\end{table}

\begin{table}[p]
\centering
\scriptsize
\setlength{\tabcolsep}{3pt}
\caption{Llama-1B prefix results (\%): mean $\pm$ sample standard deviation over repeated sampling runs. }
\resizebox{\linewidth}{!}{%
\begin{tabular}{llrrrr}
\toprule
Method & Dataset & Pass@1 & Pass@32 & Pass@64 & Pass@128 \\
\midrule
\rowcolor{gray!15} CoT-SFT & GSM8K & $49.96\pm0.32$ & $86.62\pm0.48$ & $90.45\pm0.54$ & $93.18\pm0.21$ \\
\rowcolor{gray!15} CoT-SFT & GSM-Hard & $11.45\pm0.20$ & $23.68\pm0.64$ & $25.50\pm0.44$ & $27.72\pm0.19$ \\
\rowcolor{gray!15} CoT-SFT & SVAMP & $59.50\pm0.95$ & $87.97\pm0.31$ & $91.07\pm0.25$ & $93.37\pm0.40$ \\
\rowcolor{gray!15} CoT-SFT & MultiArith & $94.07\pm1.70$ & $100.00\pm0.00$ & $100.00\pm0.00$ & $100.00\pm0.00$ \\
Coconut & GSM8K & $27.52\pm0.57$ & $49.28\pm0.50$ & $53.68\pm0.47$ & $58.07\pm0.67$ \\
Coconut & GSM-Hard & $6.47\pm0.29$ & $11.95\pm0.38$ & $13.17\pm0.36$ & $14.18\pm0.55$ \\
Coconut & SVAMP & $41.87\pm0.12$ & $53.13\pm0.29$ & $54.90\pm0.53$ & $56.97\pm0.75$ \\
Coconut & MultiArith & $59.07\pm1.95$ & $85.93\pm0.85$ & $87.41\pm0.64$ & $89.44\pm0.56$ \\
CODI & GSM8K & $54.55\pm0.05$ & $69.11\pm0.38$ & $71.08\pm0.48$ & $72.37\pm0.05$ \\
CODI & GSM-Hard & $12.24\pm0.05$ & $16.41\pm0.48$ & $17.13\pm0.43$ & $17.55\pm0.16$ \\
CODI & SVAMP & $58.75\pm0.07$ & $70.70\pm0.57$ & $72.15\pm0.07$ & $73.00\pm0.00$ \\
CODI & MultiArith & $97.22\pm0.79$ & $98.33\pm0.00$ & $98.33\pm0.00$ & $98.33\pm0.00$ \\
PLaT-8 & GSM8K & $36.20\pm0.38$ & $68.73\pm0.05$ & $75.28\pm0.00$ & $80.59\pm0.32$ \\
PLaT-8 & GSM-Hard & $8.26\pm0.00$ & $16.30\pm0.54$ & $18.23\pm0.05$ & $19.75\pm0.16$ \\
PLaT-8 & SVAMP & $43.00\pm0.95$ & $66.87\pm0.42$ & $70.63\pm0.55$ & $74.47\pm0.32$ \\
PLaT-8 & MultiArith & $84.63\pm1.16$ & $96.85\pm0.85$ & $97.41\pm1.28$ & $98.70\pm0.85$ \\
\bottomrule
\end{tabular}
}
\end{table}

\begin{table}[p]
\centering
\scriptsize
\setlength{\tabcolsep}{3pt}
\caption{Llama-3B prefix results (\%): mean $\pm$ sample standard deviation over repeated sampling runs. }
\resizebox{\linewidth}{!}{%
\begin{tabular}{llrrrr}
\toprule
Method & Dataset & Pass@1 & Pass@32 & Pass@64 & Pass@128 \\
\midrule
\rowcolor{gray!15} CoT-SFT & GSM8K & $58.88\pm0.93$ & $92.90\pm0.53$ & $95.27\pm0.69$ & $96.71\pm0.32$ \\
\rowcolor{gray!15} CoT-SFT & GSM-Hard & $18.95\pm0.15$ & $38.34\pm0.27$ & $41.37\pm0.76$ & $44.12\pm1.28$ \\
\rowcolor{gray!15} CoT-SFT & SVAMP & $74.67\pm1.05$ & $95.50\pm0.17$ & $96.90\pm0.20$ & $97.83\pm0.21$ \\
\rowcolor{gray!15} CoT-SFT & MultiArith & $97.78\pm1.67$ & $100.00\pm0.00$ & $100.00\pm0.00$ & $100.00\pm0.00$ \\
Coconut & GSM8K & $28.38\pm0.83$ & $62.32\pm0.13$ & $70.18\pm0.69$ & $75.87\pm0.34$ \\
Coconut & GSM-Hard & $6.09\pm0.37$ & $15.62\pm0.20$ & $17.29\pm0.13$ & $19.13\pm0.27$ \\
Coconut & SVAMP & $34.07\pm0.57$ & $57.47\pm0.35$ & $61.73\pm0.86$ & $65.43\pm0.55$ \\
Coconut & MultiArith & $89.26\pm2.51$ & $97.04\pm0.32$ & $97.22\pm0.00$ & $97.96\pm0.32$ \\
CODI & GSM8K & $57.16\pm0.43$ & $76.72\pm0.00$ & $80.44\pm0.43$ & $83.24\pm0.32$ \\
CODI & GSM-Hard & $13.38\pm0.27$ & $18.12\pm0.43$ & $19.45\pm0.16$ & $20.89\pm0.38$ \\
CODI & SVAMP & $68.70\pm0.71$ & $80.65\pm0.35$ & $82.60\pm0.71$ & $84.60\pm0.99$ \\
CODI & MultiArith & $99.72\pm0.39$ & $100.00\pm0.00$ & $100.00\pm0.00$ & $100.00\pm0.00$ \\
PLaT-8 & GSM8K & $39.85\pm0.85$ & $73.82\pm0.42$ & $80.09\pm0.35$ & $84.79\pm0.31$ \\
PLaT-8 & GSM-Hard & $9.48\pm0.20$ & $17.66\pm0.15$ & $19.23\pm0.04$ & $20.87\pm0.12$ \\
PLaT-8 & SVAMP & $54.77\pm1.12$ & $76.00\pm0.50$ & $79.43\pm0.64$ & $82.87\pm0.74$ \\
PLaT-8 & MultiArith & $95.37\pm0.85$ & $98.89\pm0.00$ & $99.63\pm0.32$ & $99.63\pm0.32$ \\
\bottomrule
\end{tabular}
}
\end{table}

\subsection{Ablation Results by Dataset}
\label{sec:ablation-single}
All displayed entries report means $\pm$ sample standard deviations over verified sampling runs. The semantic-step row uses one latent vector per complete training step, while fixed-$K$ rows use one latent vector per $K$-token segment; the two rows therefore change both segmentation and training configuration. Semantic-step segmentation is practical when the supervision has reliable step boundaries, but is difficult to transfer to less regularly structured solutions; fixed-token segmentation avoids requiring such boundaries. The CoT-initialization row is a historical training variant and does not isolate the effect of initialization. Decoder-context variants use the PLaT-8 configuration. OpenR1-trained evaluations are separate.
\begin{table}[htbp]
\centering
\scriptsize
\setlength{\tabcolsep}{3pt}
\caption{Llama-1B GSM8K ablations (\%).}
\begin{tabular}{lrrrr}
\toprule
& Pass@1 & Pass@32 & Pass@64 & Pass@128 \\
\midrule
CoT initialization & $29.95\pm0.43$ & $67.48\pm0.86$ & $74.15\pm0.43$ & $78.01\pm0.11$ \\
semantic-step & $37.86\pm1.10$ & $69.45\pm0.46$ & $75.16\pm0.62$ & $80.82\pm0.62$ \\
PLaT-2 & $13.98\pm0.95$ & $49.43\pm1.32$ & $58.78\pm0.86$ & $68.54\pm1.19$ \\
PLaT-5 & $34.72\pm0.96$ & $69.56\pm0.27$ & $75.93\pm0.48$ & $81.80\pm0.21$ \\
PLaT-8 & $36.20\pm0.38$ & $68.73\pm0.05$ & $75.28\pm0.00$ & $80.59\pm0.32$ \\
PLaT-8, $L=2$ & $37.04\pm0.27$ & $68.69\pm0.11$ & $74.53\pm0.21$ & $79.49\pm0.27$ \\
PLaT-8, $L=3$ & $37.38\pm1.61$ & $68.80\pm0.27$ & $74.64\pm0.48$ & $79.80\pm0.48$ \\
PLaT-16 & $37.45\pm0.11$ & $65.43\pm0.54$ & $69.71\pm0.38$ & $73.84\pm0.96$ \\
PLaT-32 & $32.79\pm0.05$ & $62.85\pm0.11$ & $68.92\pm0.75$ & $73.12\pm0.91$ \\
Decoder latent history & $38.51\pm0.11$ & $67.29\pm0.70$ & $73.62\pm0.21$ & $78.85\pm0.32$ \\
Decoder question + latent history & $31.61\pm0.95$ & $67.73\pm0.99$ & $74.37\pm0.23$ & $80.29\pm0.82$ \\
\bottomrule
\end{tabular}
\end{table}
\begin{table}[htbp]
\centering
\scriptsize
\setlength{\tabcolsep}{3pt}
\caption{Llama-1B GSM-Hard ablations (\%).}
\begin{tabular}{lrrrr}
\toprule
& Pass@1 & Pass@32 & Pass@64 & Pass@128 \\
\midrule
CoT initialization & $7.25\pm0.31$ & $16.50\pm0.12$ & $18.52\pm0.55$ & $20.04\pm0.42$ \\
semantic-step & $9.00\pm0.31$ & $17.11\pm0.32$ & $18.65\pm0.35$ & $20.50\pm0.37$ \\
PLaT-2 & $2.77\pm0.27$ & $10.80\pm0.27$ & $13.68\pm0.16$ & $16.72\pm0.48$ \\
PLaT-5 & $8.04\pm0.21$ & $17.02\pm0.27$ & $18.95\pm0.21$ & $20.58\pm0.38$ \\
PLaT-8 & $8.26\pm0.00$ & $16.30\pm0.54$ & $18.23\pm0.05$ & $19.75\pm0.16$ \\
PLaT-8, $L=2$ & $8.30\pm0.59$ & $16.91\pm0.21$ & $18.57\pm0.32$ & $19.94\pm0.21$ \\
PLaT-8, $L=3$ & $9.02\pm0.21$ & $16.76\pm0.75$ & $18.50\pm0.43$ & $19.94\pm0.32$ \\
PLaT-16 & $8.53\pm0.05$ & $15.66\pm0.05$ & $16.87\pm0.27$ & $18.01\pm0.16$ \\
PLaT-32 & $7.73\pm0.00$ & $15.09\pm0.32$ & $16.60\pm0.43$ & $18.01\pm0.27$ \\
Decoder latent history & $8.53\pm0.05$ & $16.64\pm0.80$ & $18.42\pm0.54$ & $19.71\pm0.43$ \\
Decoder question + latent history & $7.05\pm0.26$ & $16.27\pm0.16$ & $18.30\pm0.31$ & $19.86\pm0.30$ \\
\bottomrule
\end{tabular}
\end{table}
\begin{table}[htbp]
\centering
\scriptsize
\setlength{\tabcolsep}{3pt}
\caption{Llama-1B SVAMP ablations (\%).}
\begin{tabular}{lrrrr}
\toprule
& Pass@1 & Pass@32 & Pass@64 & Pass@128 \\
\midrule
CoT initialization & $36.70\pm0.14$ & $65.40\pm1.70$ & $69.55\pm0.07$ & $72.35\pm0.07$ \\
semantic-step & $44.20\pm0.44$ & $68.77\pm0.49$ & $72.37\pm0.65$ & $75.93\pm0.84$ \\
PLaT-2 & $19.50\pm0.14$ & $50.80\pm0.71$ & $57.90\pm0.85$ & $65.15\pm0.64$ \\
PLaT-5 & $45.05\pm0.21$ & $68.70\pm0.42$ & $73.15\pm0.07$ & $77.45\pm0.21$ \\
PLaT-8 & $43.00\pm0.95$ & $66.87\pm0.42$ & $70.63\pm0.55$ & $74.47\pm0.32$ \\
PLaT-8, $L=2$ & $45.10\pm0.28$ & $67.45\pm0.07$ & $71.75\pm0.21$ & $75.50\pm0.42$ \\
PLaT-8, $L=3$ & $47.10\pm0.00$ & $67.35\pm0.49$ & $70.50\pm1.27$ & $73.70\pm0.71$ \\
PLaT-16 & $42.70\pm0.57$ & $61.20\pm0.00$ & $64.35\pm0.21$ & $67.15\pm0.07$ \\
PLaT-32 & $40.65\pm0.49$ & $56.65\pm0.21$ & $59.25\pm0.07$ & $62.40\pm0.28$ \\
Decoder latent history & $45.65\pm0.07$ & $66.80\pm0.57$ & $70.35\pm0.07$ & $73.45\pm0.49$ \\
Decoder question + latent history & $33.60\pm0.95$ & $61.73\pm0.68$ & $66.53\pm0.55$ & $71.23\pm0.81$ \\
\bottomrule
\end{tabular}
\end{table}
\begin{table}[htbp]
\centering
\scriptsize
\setlength{\tabcolsep}{3pt}
\caption{Llama-1B MultiArith ablations (\%).}
\begin{tabular}{lrrrr}
\toprule
& Pass@1 & Pass@32 & Pass@64 & Pass@128 \\
\midrule
CoT initialization & $72.50\pm1.18$ & $93.61\pm0.39$ & $95.56\pm0.00$ & $95.83\pm0.39$ \\
semantic-step & $88.89\pm1.67$ & $95.74\pm0.64$ & $96.67\pm0.56$ & $96.85\pm0.64$ \\
PLaT-2 & $27.50\pm0.39$ & $86.11\pm0.79$ & $91.94\pm1.96$ & $95.00\pm0.00$ \\
PLaT-5 & $78.61\pm0.39$ & $95.83\pm0.39$ & $97.50\pm0.39$ & $98.33\pm0.00$ \\
PLaT-8 & $84.63\pm1.16$ & $96.85\pm0.85$ & $97.41\pm1.28$ & $98.70\pm0.85$ \\
PLaT-8, $L=2$ & $93.61\pm0.40$ & $98.89\pm0.78$ & $99.44\pm0.00$ & $99.44\pm0.00$ \\
PLaT-8, $L=3$ & $96.94\pm1.96$ & $99.44\pm0.00$ & $99.44\pm0.00$ & $99.72\pm0.39$ \\
PLaT-16 & $84.17\pm0.39$ & $95.00\pm0.79$ & $96.11\pm0.00$ & $96.94\pm0.39$ \\
PLaT-32 & $60.28\pm4.32$ & $82.22\pm0.79$ & $84.72\pm0.39$ & $87.78\pm0.00$ \\
Decoder latent history & $93.06\pm0.39$ & $98.06\pm0.39$ & $98.06\pm0.39$ & $99.17\pm0.39$ \\
Decoder question + latent history & $60.93\pm1.16$ & $91.67\pm1.11$ & $93.89\pm1.11$ & $96.30\pm0.32$ \\
\bottomrule
\end{tabular}
\end{table}

\subsection{Agreement of LLM Judgments}
\label{sec:appendix-agree-judge}
We drew 100 unique chunks per method without replacement using seed 20260915, independently of GPT tags.
The eligible pools contained 1,725 PLaT records and 3,951 CoT records.
We excluded 28 and 119 records, respectively, that lacked complete Boolean judgment fields or stored neighboring context.
The earlier 31-item stratified pilot is separate from this 200-item review.

All 200 human labels are valid or invalid (181 valid, 19 invalid).
The reviewer could see the GPT tags and explanations.
The resulting agreement therefore characterizes assisted review and cannot establish independent inter-rater reliability.
The mapping is invalid when \texttt{malformed=true}, valid when \texttt{valid\_local=true} and \texttt{malformed=false}, and otherwise ambiguous.
The \texttt{incomplete} tag alone does not imply invalidity.

\begin{table}[ht]
\centering
\caption{Human versus GPT labels on all 200 reviewed items. Columns are GPT labels.}
\label{appendix_tab:human-llm-agreement}
\begin{tabular}{lrrr}
\toprule
Human label & Valid & Invalid & Ambiguous \\
\midrule
Valid & 170 & 9 & 2 \\
Invalid & 5 & 14 & 0 \\
\bottomrule
\end{tabular}
\end{table}
With valid as the positive class, there are five false positives and nine false negatives, plus two GPT abstentions on human-valid items.
\begin{table}[h]
    \centering
    \caption{\textbf{Human-LLM Agreement Metrics.}}
    \begin{tabular}{lcc}
    \toprule
        Metric & Cohen's Kappa ($\kappa$) & Accuracy \\
        \midrule
        Value & 0.5943 & 0.9200 \\
        \bottomrule
    \end{tabular}
    \label{appendix_tab:human-llm-agreement-metrics}
\end{table}
Agreement is 184/200 (92.0\%).
Cohen's $\kappa=0.5943$ is computed over valid, invalid and ambiguous categories; it is not a binary calculation that discards abstentions.
PLaT has agreement 97.0\% and $\kappa=0.7845$; CoT-SFT has agreement 87.0\% and $\kappa=0.4821$.
These labels do not establish correctness of every occurrence of a repeated chunk.

\begin{figure}[htbp]
\centering
    \includegraphics[width=0.72\linewidth]{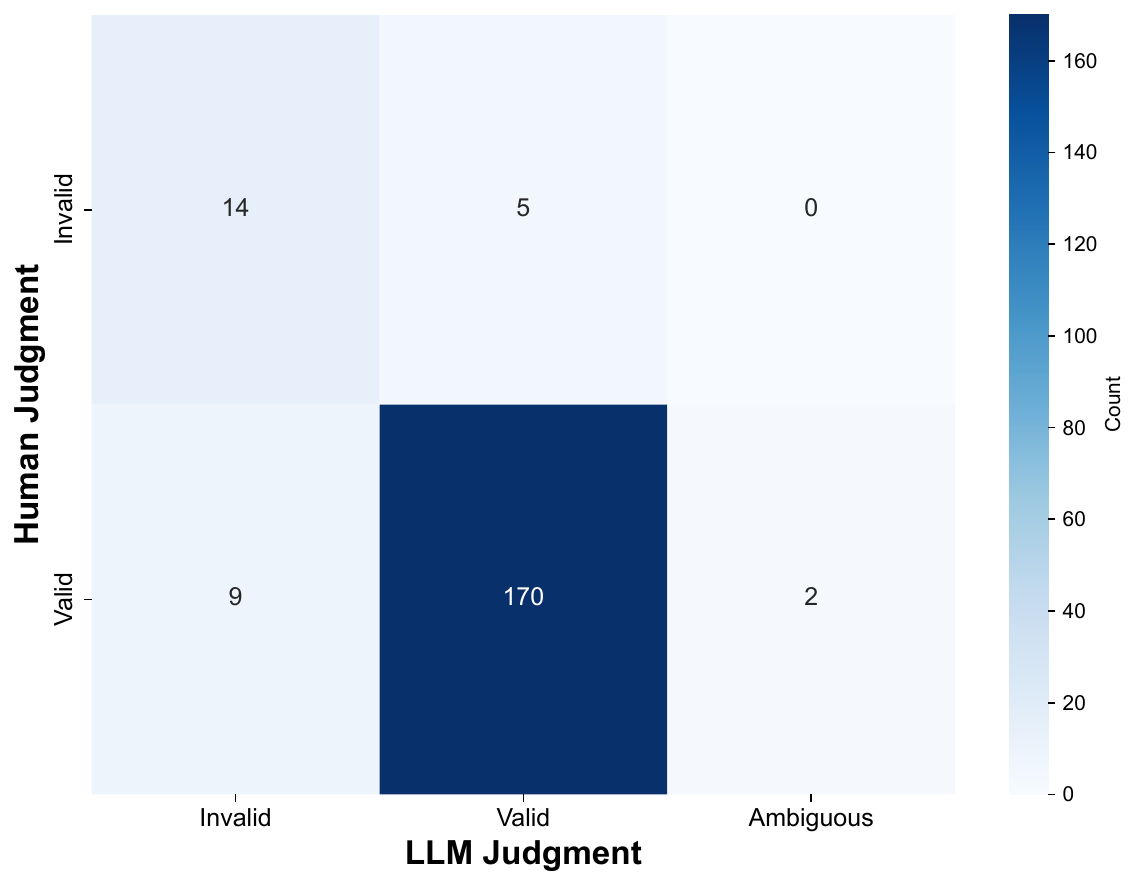}
    \caption{\textbf{Confusion matrix of Human-LLM agreement.} All 200 reviewed items are included. The two ambiguous LLM judgments count as disagreements.}
    \label{appendix_fig:human-llm-agreement-matrix}

\end{figure}

\subsection{Case Study}

\begin{figure}[tb]
    \centering
    \includegraphics[width=0.9\linewidth]{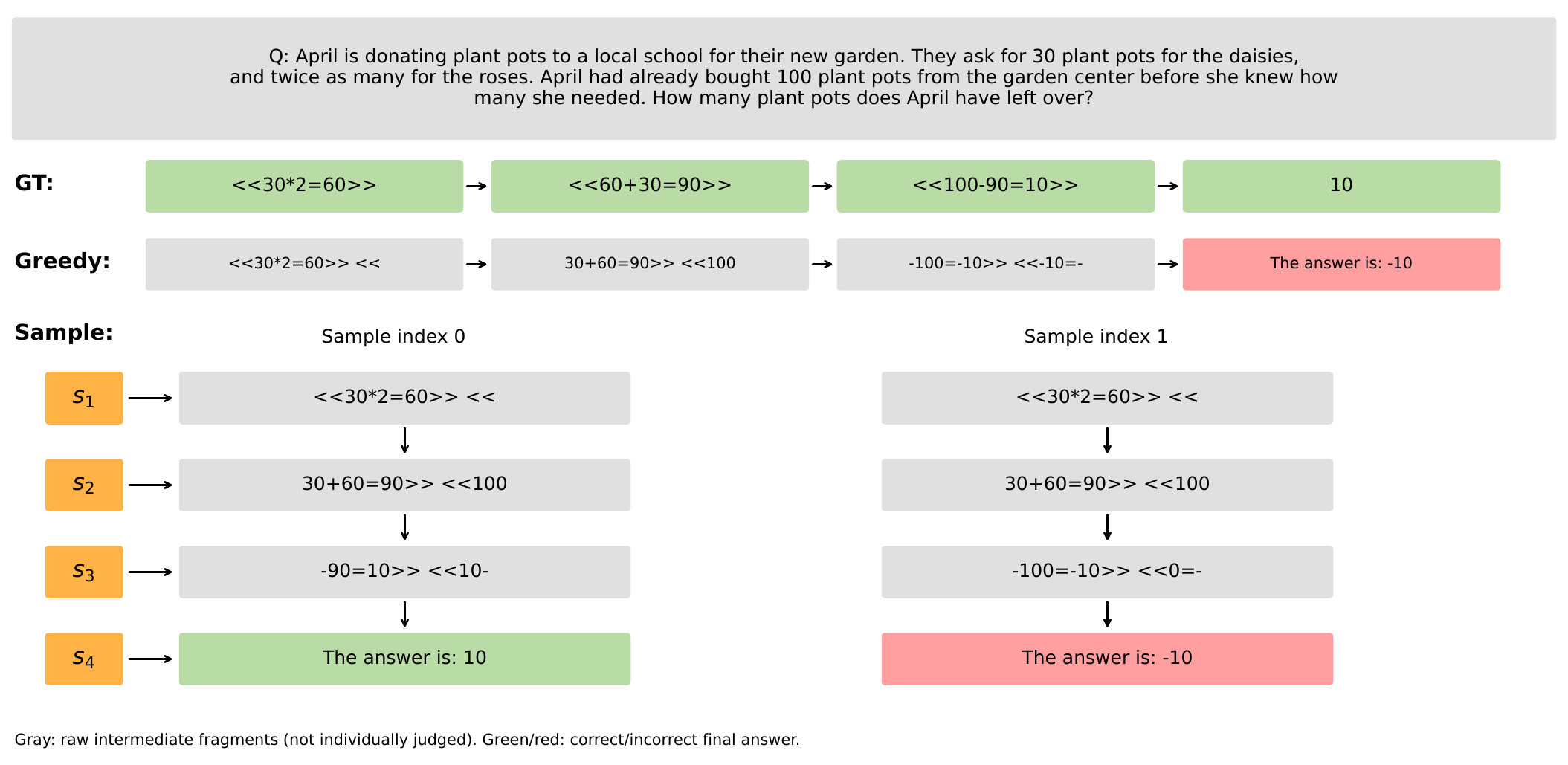}
    \caption{\textbf{Visualization of PLaT's reasoning process.} \textit{GT}, \textit{Deterministic}, and \textit{Sample} are the ground-truth reasoning steps in the dataset, the deterministic readouts from each group of states, and sampled readouts from each group of states, respectively. This example is drawn from the GSM8K test set, with two stored sampled trajectories. State labels are one-based. Gray boxes preserve raw intermediate fragments without individual validity labels; green and red final-answer boxes indicate correctness and incorrectness. Arrows connect consecutive readouts within each stored trajectory.}
    \label{fig:case_study}
\end{figure}
The deterministic answer is $-10$, while sample 0 gives the reference answer $10$ at the same answer-start state. This illustrates variation in sampled readouts, without establishing the validity of every intermediate fragment or latent superposition.

\subsection{Entropy Analysis}
\label{sec:appendix-entropy}
We evaluate the fixed-token PLaT model on all 1,319 GSM8K test questions using deterministic generation for this entropy analysis. Only question text is provided to the model.

To investigate the internal decision-making process, we measure the Shannon entropy of the first-token distribution decoded from each generated latent state. 
For a given latent state $\mathbf{S}_k$ at reasoning step $k$, let $P_{\mathcal{D}}(v \mid \mathbf{S}_k)$ denote the probability of token $v$ being the first token generated by the Decoder. 
The reasoning entropy of PLaT $H(\mathbf{S}_k)$ is defined as:
\begin{equation}
H(\mathbf{S}_k) = - \sum_{v \in \mathcal{V}} P_{\mathcal{D}}(v \mid \mathbf{S}_k) \log P_{\mathcal{D}}(v \mid \mathbf{S}_k)
\end{equation}
where $\mathcal{V}$ is the vocabulary and $\mathcal{D}$ is the Decoder. 
In our analysis, we normalize the reasoning progress of each sample to $[0, 100\%]$ to aggregate samples with varying lengths. 

\begin{figure}[htbp]
\centering
    \includegraphics[width=\linewidth]{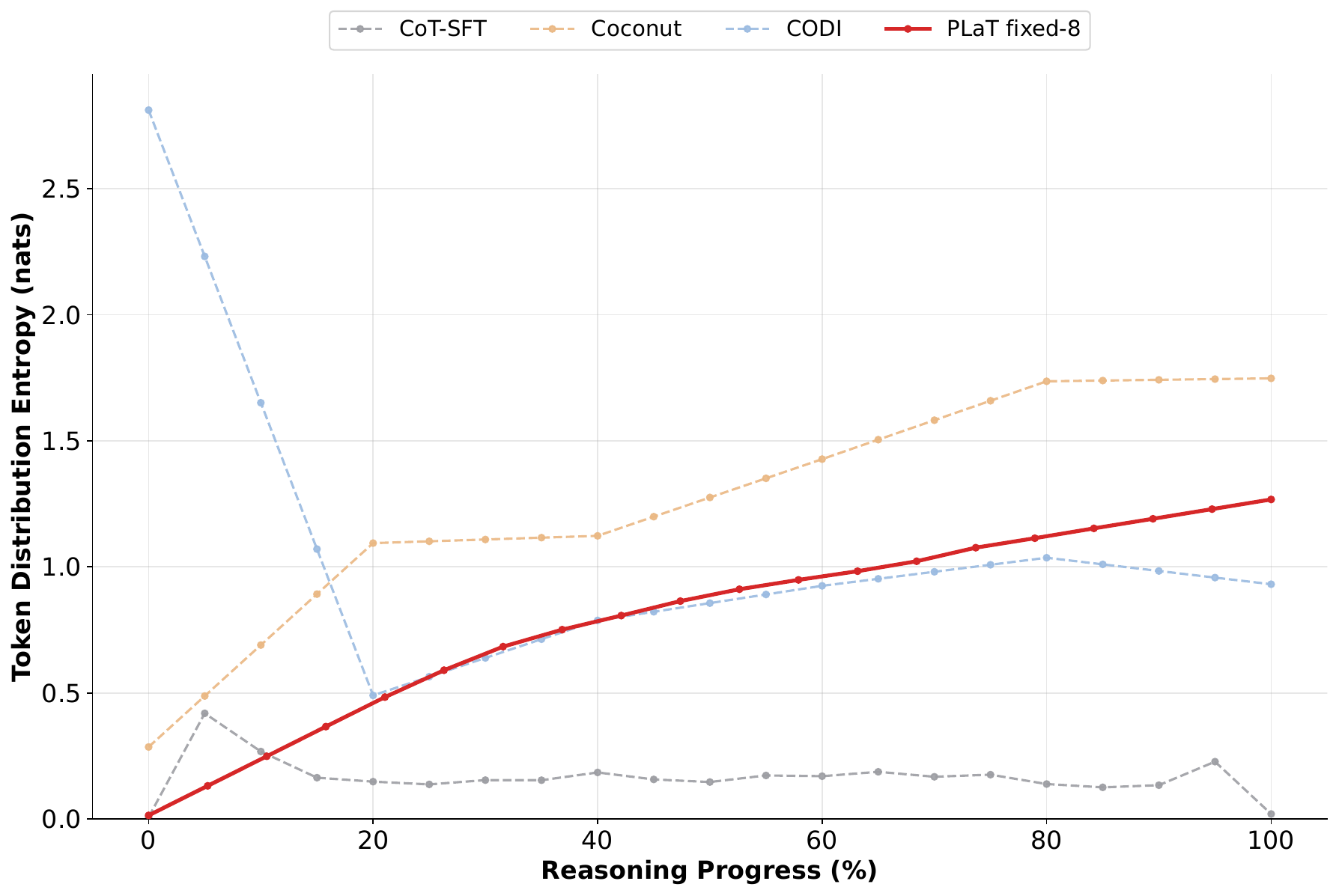}
    \caption{\textbf{Evolution of Token Distribution Entropy over Normalized Reasoning Progress.} Updated PLaT-8 first-token decoder entropy (1,318 of 1,319 questions with retained reasoning states, solid red) with original CoT-SFT, Coconut, and CODI measurements (1,319 questions each, dashed). Each question contributes equally after interpolation (20 progress points for PLaT; 21 for the original baselines). Entropy is measured in nats; baseline model outputs and method-specific probes are retained from the original analysis.}
    \label{appendix_fig:entropy}

\end{figure}

Appendix Figure \ref{appendix_fig:entropy} reports the mean entropy in nats, computed before sampling transformations. We exclude the state in which the answer prefix is detected and all subsequent states. All 1,319 generations contain a detected answer prefix. One question has the answer prefix detected in its first state and therefore has no retained reasoning state; it is excluded from the entropy average. The remaining 1,318 sequences contain 1--8 states per question; each is linearly interpolated to 20 normalized progress points (constant for a single state). These points do not represent 20 independent reasoning steps. For reference, we retain the original CoT-SFT, Coconut, and CODI measurements (1,319 questions per method); these baselines have not been rerun. To reproduce the original baseline curves, we remove the final entropy value from each cached sequence and interpolate to 21 progress points. Their original method-specific probes are retained: CoT-SFT uses generated-token distributions, while Coconut and CODI use latent-step vocabulary distributions. Accordingly, the curves are descriptive references rather than a matched-model comparison.

\subsection{Detailed Analysis of Hyperparameters}
\label{sec:appendix-abla-hyper}
We report the single-run sensitivity plots for token-group size $K$ and latent vectors per segment $L$.

\begin{figure}[htbp]
\centering
\includegraphics[width=\linewidth]{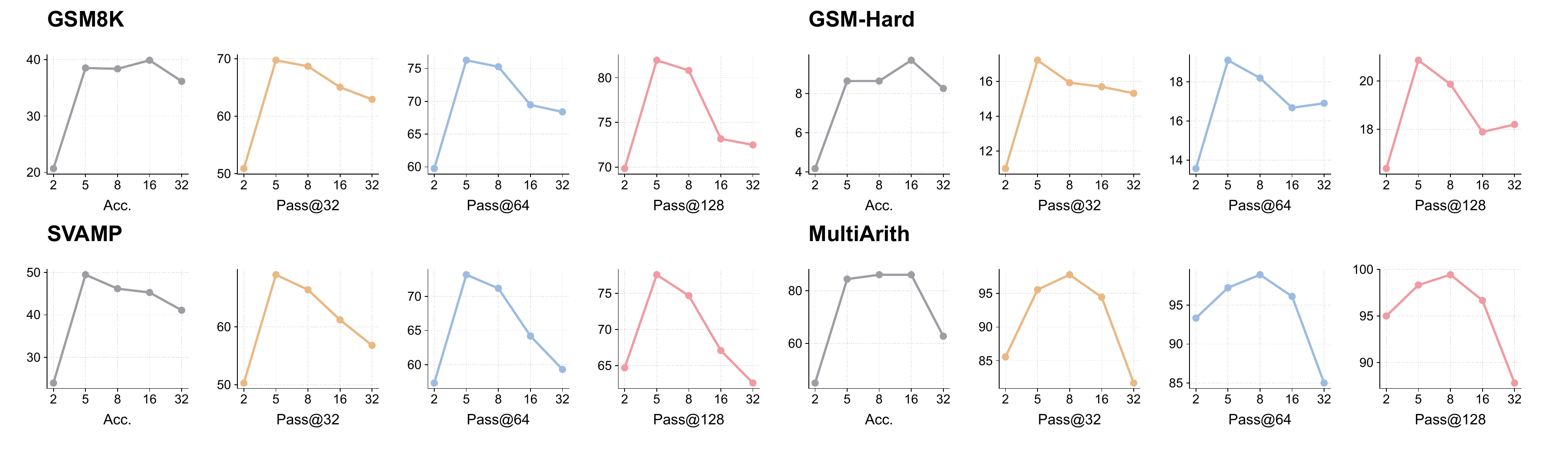}
\caption{Sensitivity to token-group size $K$ with one latent vector per segment. Single-run results on four datasets; no mean or standard deviation across repeated runs is reported.}
\label{appendix_fig:hyper-token-size}
\end{figure}

\paragraph{Token-group Size ($K$).}
Figure~\ref{appendix_fig:hyper-token-size} reports $K=2,5,8,16,32$ with $L=1$.

\paragraph{Sensitivity to Latent Vectors per Segment ($L$).}
The sweep varies $L=1,2,3$ at fixed $K=8$.
\begin{figure}[htbp]
\centering
\includegraphics[width=\linewidth]{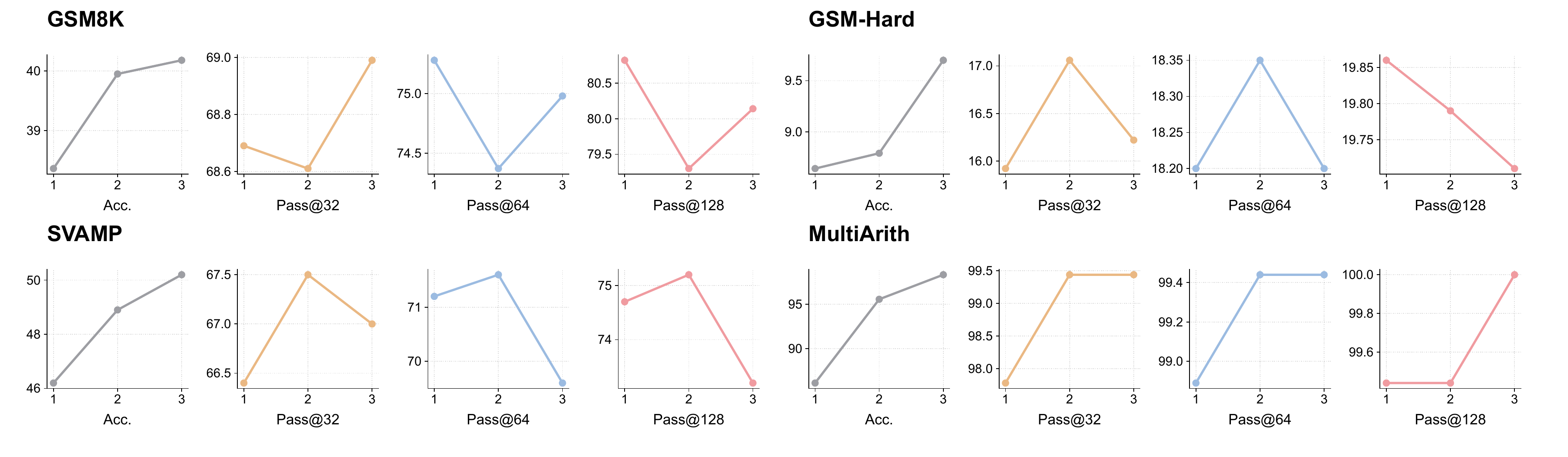}
\caption{Impact of latent vectors per segment $L$ at fixed token-group size $K=8$. Single-run results on four datasets, without repeat-run error bars.}
\label{appendix_fig:hyper-latent-num}
\end{figure}

\section{Prompts for LLM Judgments}
\label{sec:appendix-prompt}
\label{sec:appendix-prompt-validate}
The current evaluator receives the problem/reference reasoning, candidate chunk, and stored previous and next readouts.
It first inspects the local window for equations split across boundaries.
A plausible arithmetic fragment with no explicit complete arithmetic error may be marked locally valid even when truncated.
Only genuinely nonsensical content or an explicitly incorrect complete equality is marked malformed.
Reference reasoning is contextual information, not a requirement to follow the reference solution's ordering or path.
The output fields are \texttt{coherent}, \texttt{valid\_local}, \texttt{incomplete}, \texttt{malformed}, and \texttt{reason}.
This permissive local criterion differs from proving the correctness of a complete reasoning step.
No semantic clustering is used in the current exact-text branch counts.
\begin{tcblisting}{
  title=Prompt for Validating Equations,
  colback=white,
  listing only,
  breakable,
  listing options={
    basicstyle=\normalfont\small,
    breaklines=true,
    breakautoindent=false,
    breakindent=0pt,
    columns=fullflexible
  }
}
You evaluate one candidate fixed-token readout from mathematical reasoning.
The candidate is a short token window, not necessarily a complete sentence or equation.
The symbols << and >> are formatting/chunk-boundary markers: ignore them completely.
Gold reasoning steps and gold fixed-token chunks are reference only. They explain the
problem and typical operations, but do NOT require the candidate to match the gold
solution, prefix, ordering, or arithmetic continuity.

The user message includes the previous and next readout from the SAME generated
trajectory when available. These are context, not additional candidate text.
First inspect the concatenated local window to determine whether an apparent
equation is split across chunk boundaries. Then judge the candidate chunk:
- If it contains a plausible arithmetic/semantic fragment and no explicit complete
  arithmetic error, set valid_local=true.
- A fragment may begin/end mid-number, operator, or equation; missing context is not
  evidence of invalidity. Set incomplete=true when visibly truncated, but keep
  valid_local=true if the visible operation is plausible.
- Do not mark a candidate malformed merely because it looks like `a=b` in isolation
  when the neighboring readouts show it is the suffix/prefix of a larger expression
  such as `c+a=b`.
- Set malformed=true only for genuinely nonsensical text or an explicitly incorrect
  complete equality (e.g. 3*43=128). Do not call boundary markers, dangling operators,
  or missing operands malformed.
Return JSON only with keys coherent, valid_local, incomplete, malformed, reason.
\end{tcblisting}

\section{Limitations and Future Work}

Candidate coverage and textual inspectability have distinct evidential limits. Higher Pass@$k$ does not directly establish better final-answer selection, and locally valid readouts do not establish complete-chain consistency or causal faithfulness to the Planner's computation. Evaluating a common answer selector and testing how latent-state interventions affect readouts and final answers would address these questions.

The architecture permits a variable number of latent groups, but the current experiments do not establish how performance scales with latent computation. The single-run sensitivity results vary with token-group size $K$ and latent-vector count $L$, without establishing monotonic improvement or isolating the effects of optimization, representation capacity, and readout configuration. Controlled studies are needed to separate these factors. The efficiency measurements cover GPT-2 greedy inference; costs for other backbones and the multi-sample regime remain to be characterized.

Our evaluation covers mathematical reasoning; transfer to less structured tasks remains untested. The reported sampling repeats quantify variation in evaluation with fixed trained weights, rather than variation across independent training runs. Broader task coverage and independently trained repetitions would help assess the generality and stability of the observed trade-offs.

\end{document}